# Application of unsupervised artificial neural network (ANN) self-organizing map (SOM) in identifying main car sales factors

Mazyar Taghavi[a]*

a: Payame Noor University, Tehran, Iran

*: Corresponding author, Email address: mazyartaghavi@gmail.com

## Abstract:

Factors which attract customers and persuade them to buy new car are various regarding different consumer tastes. There are some methods to extract pattern form mass data. In this case we firstly asked passenger car marketing experts to rank more important factors which affect customer decision making behavior using fuzzy Delphi technique, then we provided a sample set from questionnaires and tried to apply a useful artificial neural network method called self-organizing map (SOM) to find out which factors have more effect on Iranian customer's buying decision making. Fuzzy tools were applied to adjust the study to be more real. MATLAB software was used for developing and training network. Results report four factors are more important rather than the others. Results are rather different from marketing expert rankings. Such results would help manufacturers to focus on more important factors and increase company sales level.

1. Introduction

   **Car sales factors**

Passenger car industry is a promising and attractive market for investors. Market specialists always try to increase sales to satisfy strategic beneficiaries. Sales growth in car industry depends on many factors. Some are recognized by professionals and some others are intangible. There are many attributes must be considered as empowering sales factors, and every company need to know which ones are more important in current situation in order to concentrate on improving its capabilities around those main factors. Numerous studies have been conducted in Iran about passenger car sales factors. Atafar et al (2008) studied main factors in car sales and found out quality and price is the most important ones. Khodadad hoseini et al (2003) applied an AHP technique and found the quality as the most effective subject in car sales in Iran. Abdolvand et al (2008) conducted a research and identified appearance as the most important factor. Yet, none of studies have used neural network for extracting features among mass dada.

2. Literature review

2.1. Artificial Intelligence

Artificial Intelligence (AI) is a branch of computer sciences including designing intelligent computer systems. That is, such systems exhibit characteristics we recognize them as intelligent in human behavior (Avron Barr & Feigenbau, 1981). AI is a branch of computer sciences which is related to automating intelligent behavior (Luger & Stubbfiel, 1993). We can consider AI as modeling, applying, and running intelligent theories such as expert systems, neural networks, fuzzy logic, and cellular automata. Among them, neural networks, fuzzy logic, and genetic algorithm are recognized as soft computing. Fuzzy Logic, Neural Networks, and Genetic Algorithms, are such techniques used in modeling intelligence (Figure 1).

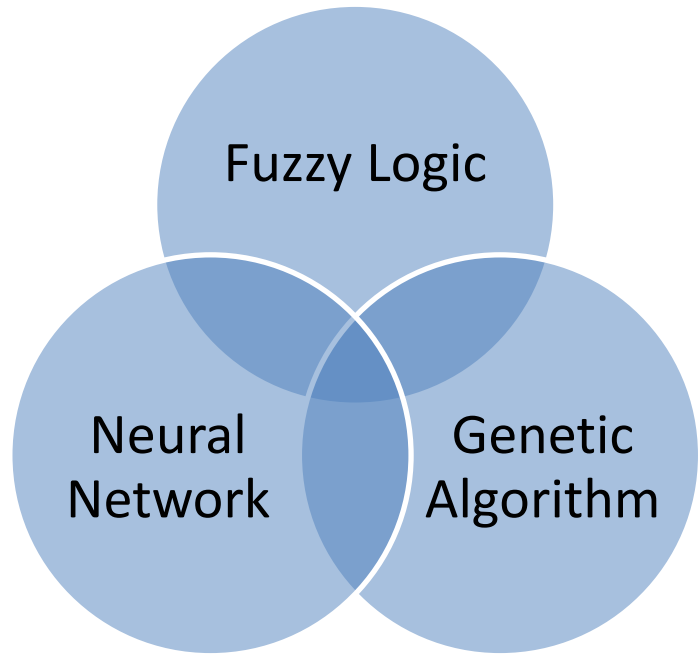

Figure 1 - Artificial Intelligence

### 2.2. Artificial Neural Networks

Artificial Neural networks are simplified models of human neural system which have acquired their instructions from calculation method in human brain. Generally, neural networks are set of linked networks with a lot of processing elements called neurons. Neural network can perform parallel. They have abilities such as mapping or modeling, generality, strength, fault tolerance, and very fast parallel processing. They are successfully used in recognition, image processing, data comparison, prediction, and somehow optimization.

Neural networks use different learning mechanisms such as supervised learning and unsupervised learning. In supervised learning method it is assumed that there is a learner during training process. That is, network can decrease error between goal output, presented by learner, and calculated output. But in unsupervised learning there is no instructor, so the network itself organizes input data and tries to learn.

### 2.3 Self-Organizing Map

The Self-Organizing Map (SOM) is a fairly well-known neural network and indeed one of the most popular unsupervised learning algorithms invented by Finnish Professor Teuvo Kohonen in the early 1980s. The maps comprehensively visualize natural groupings and relationships in the data and have been successfully applied in a broad spectrum of research areas ranging from speech recognition to financial analysis. Self-organizing networks have the ability to learn and detect regularities and correlations in the inputs, and predict responses from input data (Westerlund, 2005). A Self Organizing Map (SOM) is a single layer neural network, where neurons are set along an n-dimensional grid; typical applications assume a 2-dimensions rectangular grid. Each neuron has as many components as the input patterns: mathematically this implies that both neuron and inputs are vectors embedded in the same space. Training a SOM requires a number of steps to be performed in a sequential way. Unlike many other types of NNs, the SOM doesn't need a target output to be specified. Instead, where the node weights match the input vector, that area of the lattice is selectively optimized to more closely resemble the data for class of the input vector is a member of (Dragomir et al, 2014).

For a generic input pattern x we will have:

1. To evaluate the distance between x and each neuron of the SOM;

2. To select the neuron (node) with the smallest distance from x. We will call it winner neuron or Best Matching Unit (BMU). The algorithm used to find the "winning" neuron is a Euclidean distance calculation. That is, the "winning" neuron n of dimension d for sample input v (also of dimension d) would be the one which minimized the equation:

$$\sqrt{\sum_{i=0}^{d} (n_i - v_i)^2}$$

(2.1)

3. To correct the position of each node according to the results of Step 2, in order to preserve the network topology. The amount to adjust each "neighbor" by is determined by the following formula:

$$n_i(t + 1) = n_i(t) + h * [v(t) - n_i(t)]$$

$n(t)$ = weight vector of neuron i at regression step t
$v(t)$ = input vector at regression step t
$h$ = neighborhood function

(2.2)

Weight vectors in the map in the neighborhood of the BMU are updated so that they move closer to the sample vector (see Figure 2):

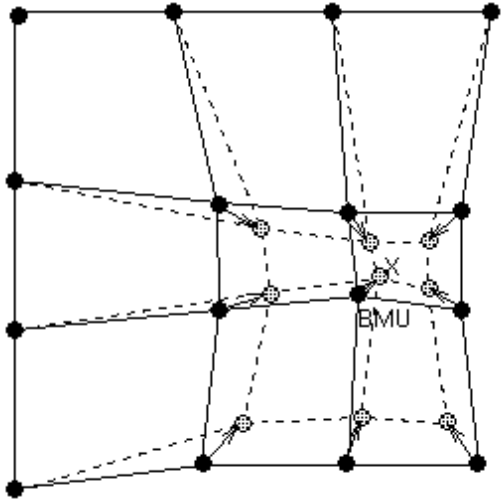

Figure 2– Self-organizing feature lattice

## 2.4. Fuzzy Logic and Fuzzy numbers

Fuzzy set theory was introduced by Lotfi Askarzadeh (Zadeh) in 1965. Fuzzy logic could be described in mathematics by allocating value to each possible affair in reference set. These values show membership function in fuzzy set. This membership function shows correspondence, similarity, or adaptive capability of that element to the concept which is presented in fuzzy set. In other words, fuzzy sets provide flexible membership function for each member of the set. In classic sets theory a member belongs to a set or not. So, they call such sets "crisp sets". But in fuzzy sets many membership functions (between zero and one) is allowed. Therefore, membership function $\mu_A(x)$ in fuzzy set writes each member of fuzzy reference set into the interval [0 1], just like a pattern. A fuzzy number (see figure 3) is a generalization of a regular, real number in the sense that it does not refer to one single value but rather to a connected set of possible values, where each possible value has its own weight between 0 and 1. Triangular and trapezoid fuzzy numbers are defined as following equations:

$$\mu_{f(x)} = \begin{cases} \frac{x-l}{m-l} & l < x < m \\ \frac{u-x}{u-m} & m < x < u \\ 0 & otherwise \end{cases} \quad (2.3)$$

$$\mu_{M(x)} = \begin{cases} \dfrac{x-l}{m-l} & l < x < m \\ \\ 1 & m < x < n \\ \\ \dfrac{x-u}{n-u} & n < x < u \\ \\ 0 & otherwise \end{cases} \qquad (2.4)$$

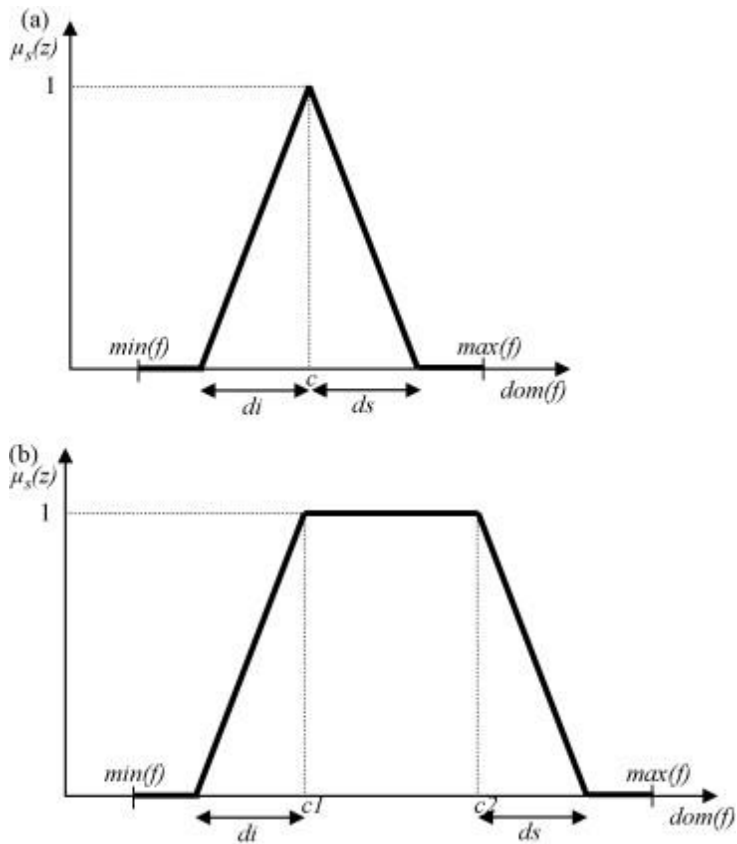

Figure 3 - Fuzzy triangular and trapezoidal number

### 2.5. Fuzzy Delphi method

The Delphi method originated in a series of studies that the RAND Corporation conducted in the 1950s (Okoli & Pawlowski, 2004). The objective was to develop a technique to obtain the most reliable consensus of a group of experts (Dalkey & Helmer, 1963). Although, in this research we could select the critical variables through the traditional statistical analysis methods, Delphi method was used as a stronger methodology to achieve consensus of our market managers on final variables set. Okoli and Pawlowski (2004) compared and contrasted the strengths and weaknesses of a Delphi study versus the traditional survey approach as a research strategy (Okoli & Pawlowski, 2004).

### 3. Empirical study

In this research main factors which affect buying behaviors in Iran car market was studied. Our objective is to recognize which factors create more motivation in customers to choose and buy an Iran-made car.

Our methodology and steps to do the research is explained as following.

### 3.1 Research methodology

Figure 4 demonstrates our work procedure. The research has been conducted in three stages.

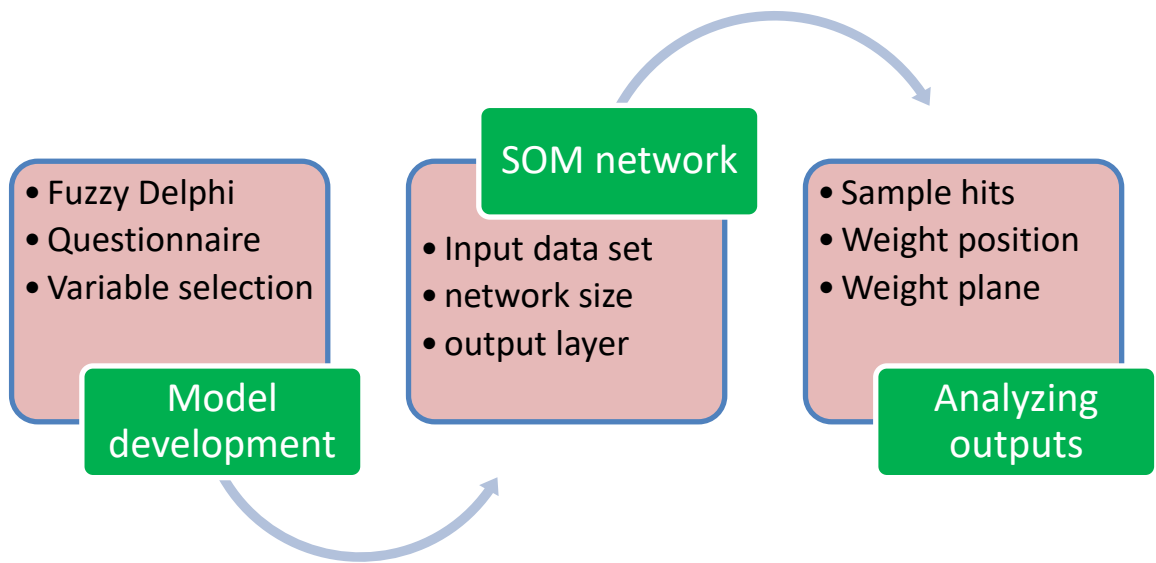

Figure 4 - Conceptual model

**Stage 1**: Model development

a. Defining the problem, specifying research goals, scope, and methodology.

b. Providing a list of main factors.

c. Preparing proper questionnaire and providing input items.

**Stage 2**: SOM network development

a. Model development, network construction, and parameters configuration.

b. Making the data set using obtained data from Iran khodro Company customers by means of an electronic survey and customers' data available at the database of sales and after sales service system.

c. Data preprocessing by standardizing and normalizing the data set.

d. Setting initial weights and training the network using data set.

e. Generating variable maps.

**Stage 3**: Output analysis

a. Analyzing each input variable

b. Comparing inputs

c. Identifying important factors

### 3.2 Model development

Iran khodro Company is the largest auto maker in middle-east with more than 550000 annual productions, and produces vast variety of cars with different types. Finding the most factors which cause sales growth would help the car manufacturer to adopt best marketing strategy. Because there are many factors which increase sales amount and finding main factors among them is so hard regarding large scale data, we decided to employ fuzzy Delphi method to identify the most effective factors. Delphi method is used for the following reasons:
Firstly our study is an investigation of factors that would affect customer behavior in Iran market. This special subject requires expert people who understand the different attributes of our market structure. Thus, a Delphi method answers the research questions more appropriately. Secondly our sample size for an acceptable statistical analysis is small therefore, and also we encounter a problem which is lack of professionals in market analysis, in these conditions, Delphi technique will lead to more correct results. Finally among other group decision analysis methods Delphi is more suitable because it does not require the experts to meet physically, it save time and cost required for collecting expert's opinions. The traditional Delphi method has always suffered from low convergence expert opinions, high performance cost, and the possibility that opinion organizers may filter out particular expert opinions (Murry, Pipino, & Gigch, 1985). Thus we decided to propose the concept of integrating the traditional Delphi method and fuzzy theory to improve the vagueness and ambiguity of Delphi method. Hsu and Yang (2000) adopted triangular fuzzy number to gather expert opinions and establish Fuzzy Delphi method. The max and min values of expert opinions are taken as the two terminal points of triangular fuzzy numbers, and the geometric mean is taken as the membership degree of triangular fuzzy numbers to derive the statistically unbiased effect and avoid the impact of extreme values.

This method may create better results in criteria selection. It benefits the advantage of simplicity, and all the expert opinions can be involved in one investigation (Okoli & Pawlowski, 2004). To make triangular fuzzy number for each input variable following equations were used:

$$T_{Aj} = (L_{Aj}, M_{Aj}, U_{Aj}) \quad (3\text{-}1)$$

$$L_{Aj} = \min(X_{Aij}) \quad (3\text{-}2)$$

$$U_{Aj} = \max(X_{Aij}) \quad (3\text{-}3)$$

$$M_{Aj} = \sqrt[n]{\prod_{i=1}^{n} X_{Aij}} \quad (3\text{-}4)$$

Eq. (3-1) represents triangular fuzzy number of $j^{th}$ variable. $L_{Aj}$ is the minimum value of $A_j$ that is evaluated by experts. $U_{Aj}$ is the maximum value of $A_j$ that is evaluated by experts. $M_{Aj}$ is geometric mean of expert evaluation from $A_j$, and $X_{Aij}$ is the evaluation of $i^{th}$ expert from $i^{th}$ variable. Geometric mean of each triangular number relates to consensus of experts on the value of variable. If the geometric mean is more than a predefined threshold the variable will be selected. For the threshold value r, the 80/20 rule was adopted (Kuo & Chen, 2007) with r set as 4. This indicated that among the factors for selection, ''20% of the factors account for an 80% degree of importance of all the factors''.

The Fuzzy Delphi questionnaire in our study consists of a list of 10 variables frequently mentioned in sales factor literature. The questionnaire was distributed to fifteen managers of the sales, marketing and after sales service units. The respondents were asked to indicate to what extend each variable influences customer buying behavior, according to their contact with automobile customers.

Considering fuzzy numbers, ten variables identified as the most effective factors on the consensus of market experts. Table 1 presents the most effective variables in car sales in Iran khodro Company context.

| Variable | Quality | Price | After sales service | Representative quantiti | Leasing sales accessibi | Foreign parts | Technology | Appearance | Utility | Fuel consumption |
|---|---|---|---|---|---|---|---|---|---|---|
| Rank | 2 | 1 | 6 | 10 | 3 | 4 | 9 | 7 | 5 | 8 |

Table 1 - Most effective variables in car sales

### 3.3 SOM network development
3.3.1 Dataset

Factors identified from fuzzy Delphi method need to be short listed to provide less alternatives for car manufacturer to focus on. To achieve our goal we used data acquired from 98 useable questionnaires. The questionnaires were pre-tested by interviewing a few customers to avoid probable misunderstanding, and then were distributed to research community. They contain questions about 10 factors which affect most on buying decision making. The respondents were asked to indicate on a fuzzy scale to what extend each variable influences their purchasing behavior. The answers could be considered as fuzzy trapezoid numbers. Just like explained procedure in section 3-2, if the geometric mean is more than a predefined threshold the variable would be selected.

3.3.2. Data processing

Scaling variables is important in the SOM network creation because the SOM algorithm applies Euclidean metric to measure distances between vectors. If one variable has values in the short range and another in long range the former will almost completely dominate the map organization because of its greater impact on the distances measured. So we normalized dataset and scaled all variables linearly so that their variances were equal to one.

3.3.3. Network construction and training

Provided data sheet in excel was used in clustering tool box of MATLB® software. This graphical user interface (GUI) gets inputs and asks for network size. You may apply any needful changes in parameters and specifications. The training dataset consisted of 98 ten-dimensional vectors each corresponding to one customer data. A 10 _ 10 sized SOM was trained with this data. Regarding our sample size, 10 variables were introduced to the software as inputs and the number of samples was 98. Network size has been adjusted to a 10 * 10 grid to provide clarity in diagrams. Therefore, the output layer of the artificial network was a 100-neuron network. Initial weights of network were randomly assigned. Sequential training algorithm and Gaussian neighborhood function were used for training process. Training was performed in two steps. In the first phase, relatively large initial learning rate and neighborhood radius were used. In the second step both learning rate and neighborhood radius was small just from the beginning. These steps are designed to tune the SOM approximately according to input data space and then they fine tune the self-organizing map. Number of epochs adjusted to 200 iterations, it means training procedure occurred 200 times and then system stopped. Training method was based on batch weigh processing and performance calculated according to mean squared error method (MSE). The mean squared error (MSE) of an estimator measures the average of the squares of the "errors", that is, the difference between the estimator and what is estimated.

3.3.4. Visualization; software outpoot

For SOM training, the weight vector associated with each neuron moves to become the center of a cluster of input vectors. In addition, neurons that are adjacent to each other in the topology should also move close to each other in the input space, therefore it is possible to visualize a high-dimensional inputs space in the two dimensions of the network topology. Variable maps reveal knowledge underlying in customers data. Elementary visualization technique was firstly introduced by Ultsch and Siemon (1990). In this method relating to each attribute value in the weight vector, a RGB (Red–Green–Blue) vector and consequently a color is considered in a way that all values can be shown in a colored spectrum from dark blue (for lowest values) to dark red (for highest values). In this way, for each attribute, the color of each neuron is identified and the map related to that attribute is obtained. In 2015 version of MATLAB software which is used in our study dark red colors represent large weights and light colors represent small weights. Black ones represent weight zero. With the attribute's maps, it is then possible to evaluate the mutual relation between them, this would be correlation test. The same color of the corresponding parts of two maps indicates the correlation of the corresponding attributes in those maps.

The intensity of color's difference or similarity between maps can show the correlation rate between two variables in different parts of the space. Investigate some of the visualization tools for the SOM:

**SOM Neighbor Distances**

To view the U-matrix see figure 5. In the figure, the blue hexagons represent the neurons. The red lines connect neighboring neurons. The colors in the regions containing the red lines indicate the distances between neurons. The darker colors represent larger distances, and the lighter colors represent smaller distances. The SOM network appears to have clustered the dataset into ten distinct groups. Clustering data is another excellent application for neural networks. This process involves grouping data by similarity.

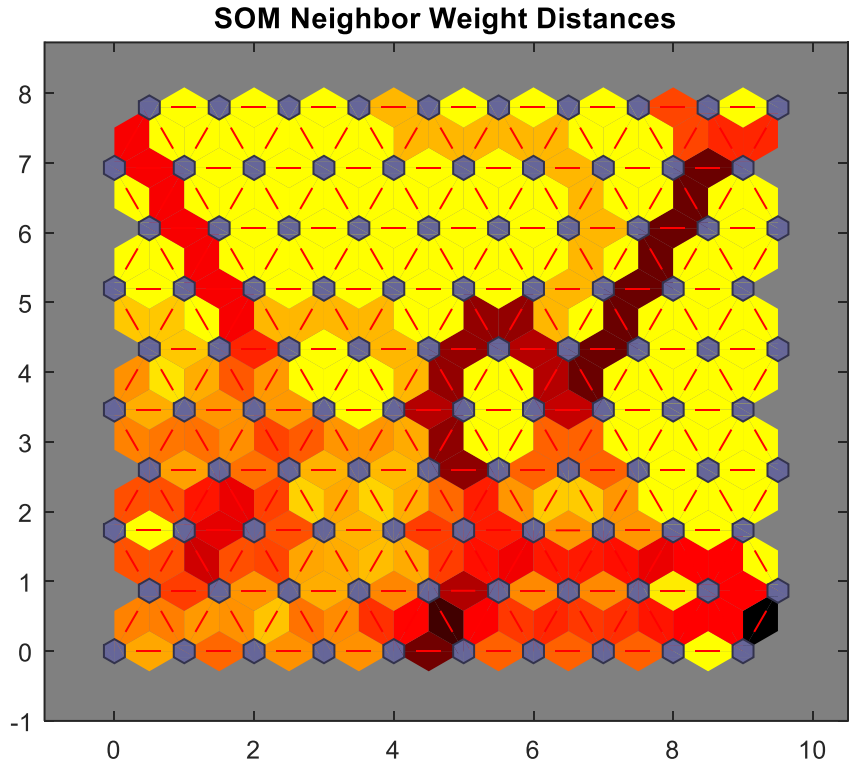

Figure 5 - The U-matrix - SOM Neighbor Distances

**Sample hits**

The default topology of the SOM is hexagonal. Figure 6 shows the neuron locations in the topology, and indicates how many of the training data are associated with each of the neurons (cluster centers). The topology is a 10-by-10 grid, so there are 100 neurons. The maximum number of hits associated with any neuron is 7. Thus, there are 7 input vectors in that cluster.
An interesting visualization way is to plot a hit histogram. SOM sample hits diagram reports number of hits which are attracted to neurons (nodes) in 100-neuron network. In this figure each hexagonal represents a neuron and each neuron is neighboring at most sex neurons. This useful figure can tell you how many data points are associated with each neuron. It is best if the data are evenly distributed across the neurons. In this example, the distribution isn't fairly even.

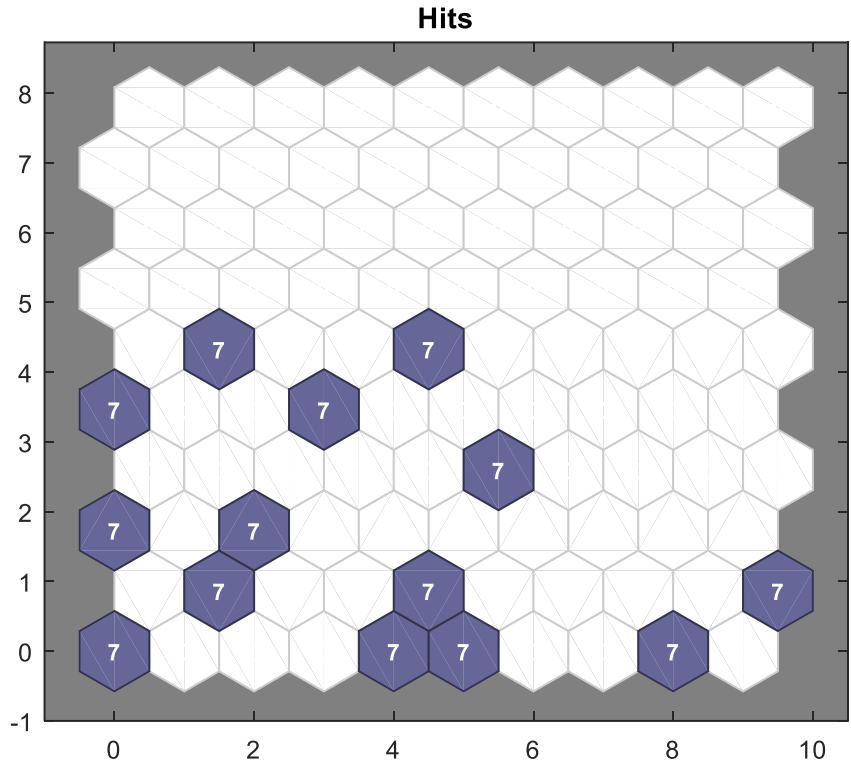

Figure 6 - SOM sample hits

**Weight position**

SOM weight positions (figure 7) shows neurons in dots and allocated weight vectors.

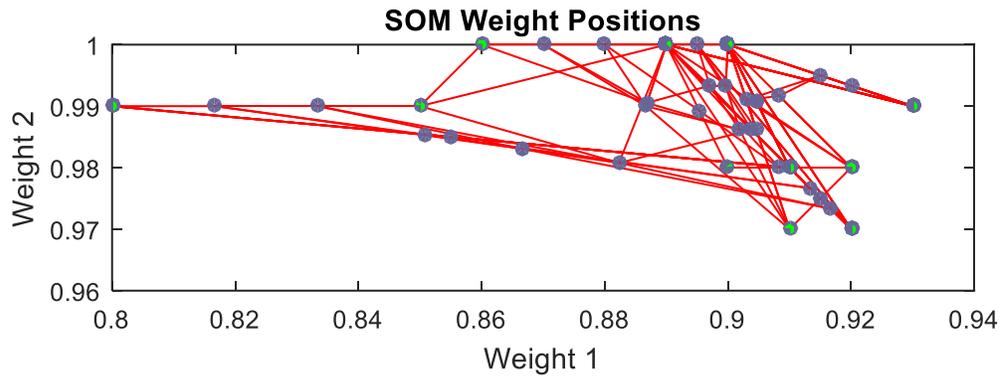

Figue 7 - weight position diagram

**Weight plane**

You can also visualize the weights themselves using the weight plane figure (see figure 8). There is a weight plane for each element of the input vector (ten, in this case). Figure 8 is called SOM input plane. It demonstrates importance of each factor. They are visualizations of the weights that connect each input to each of the neurons. Relating to each attribute's value in the weight vector, a colored vector and consequently a color is considered in a way that all values can be shown in a colored spectrum from light colors (for lowest values) to dark red (for highest values). In 2015 version of MATLAB software which is used in our study dark red colors represent large weights and light colors represent small weights. Black ones represent weight zero. If the connection patterns of two inputs are very similar, you can assume that the inputs were highly correlated.

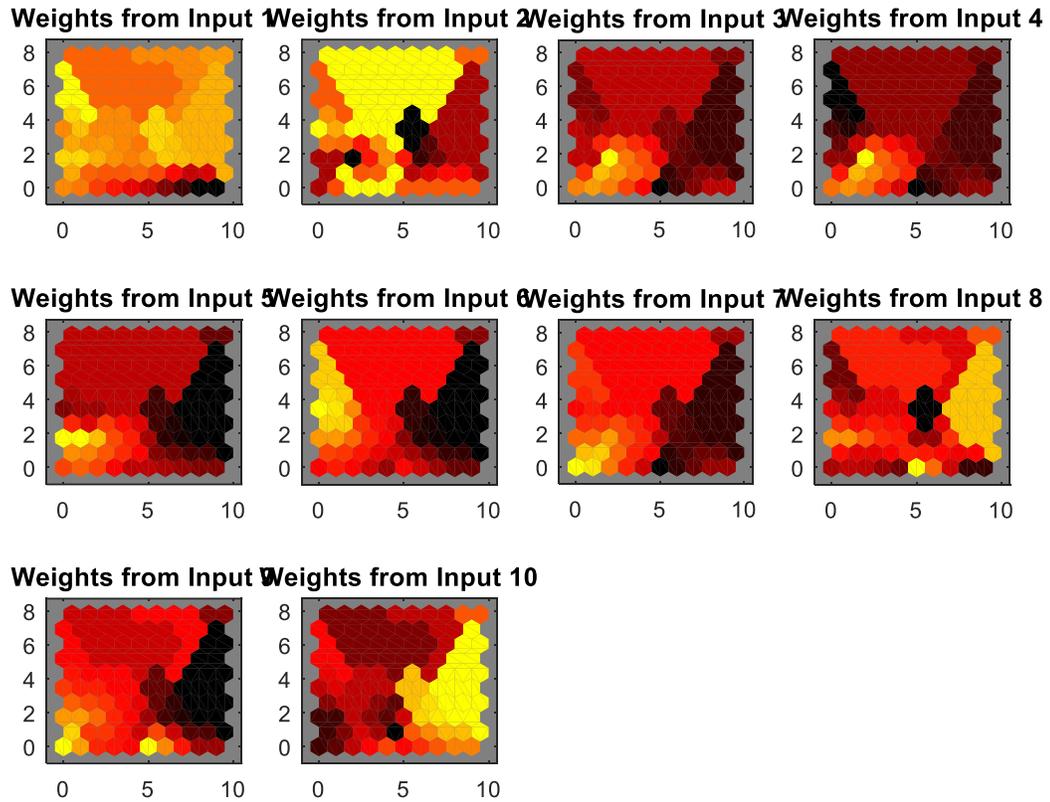

Figure 8 – Weight planes

### 3.4 Factor analysis

Some rules can be inferred from the maps. Variables 1 and 2 have more light red colors so they could be considered as high weighted neurons. Variables 9 and 7 are low weighted neurons. So, factors such as price, quality, leasing accessibility and extent of foreign parts are more important factors and variables such as utility and technology have less important rather than others.

### 4. Results

Analyzing weights for each factor and comparing to the others, the pattern embedded in the mass data is recognized. The figure 8 shows input 5, 6, 2 and 1 are most important factors among ten.

So, following main factors which affects increase in car sales amount could be identified as follow:

1. Price and 2. Quality 3.Leasing accessibility 4. Foreign parts

Also less important factors could be observed which includes utility and technology. Table 2 reports differences in results obtained from our study with the rankings already suggested by marketing experts.

| Variable | Quality | Price | After sales services | Representative quantiti | Leasing sales accessibi | Foreign parts | Technology | Appearance | Utility | Fuel consumption |
|---|---|---|---|---|---|---|---|---|---|---|
| Marketing expert rank | 2 | 1 | 6 | 10 | 3 | 4 | 9 | 7 | 5 | 8 |
| Study result rank | 2 | 1 | 6 | 8 | 3 | 4 | 9 | 7 | 10 | 5 |

Table 2 – Study results compare to marketing expert ranking

The same comparison is illustrated in Figure 9.

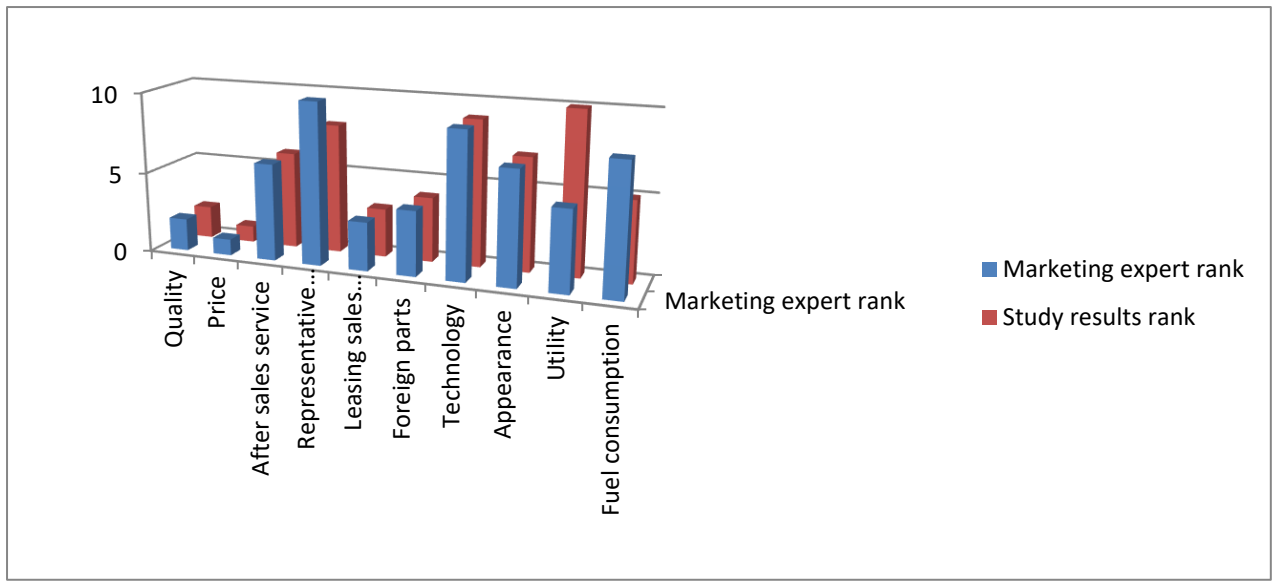

Figure 9 - Study results compare to marketing expert ranking

As we could see large conformity could be recognized in research results to the ones obtained from market experts in Iran khodro Company especially in factors which are more important. This thus could persuade marketing managers to pay more attention to these important factors in their marketing strategies. Also such results represent a new method which is more appropriate for large data.

## 5. Conclusions

An artificial neural network (NN) self-organizing map (SOM) is a data visualization technique. SOMs map multidimensional data onto lower dimensional subspaces where geometric relationships between points indicate their similarity. In this study a set of data acquired from 98 samples were considered as input layer to become resize to lower dimensional network. This helps us visualize mass data and recognize the pattern. Results show us main factors which are more important to consumers and can be used for car manufacturers.

Although there may be other methods to identify such results as well but power of artificial neural network methods mixed by fuzzy tools would increase while data size is huge.

## 6. Acknowledgement

We need to thank all questionnaire respondents and their supervisors who helped us to gather data.